\documentclass[letterpaper]{article}
\usepackage{aaai17}
\usepackage{color}

\usepackage{anyfontsize}
\usepackage{times}
\usepackage{helvet}
\usepackage{courier}
\usepackage{enumitem}
\usepackage{amssymb}
\usepackage{amsmath}

\usepackage{amsthm}
\usepackage{graphicx}
\usepackage{url}
\usepackage{multirow}

\usepackage[linesnumbered,vlined,ruled]{algorithm2e}

\usepackage{tabularx, booktabs}

\theoremstyle{plain}

\theoremstyle{definition}

\newtheorem{defn}{Definition}

\begin{document}

\title{Multi-Agent Path Finding with Delay Probabilities\thanks{Our research
    was supported by NSF under grant numbers 1409987 and 1319966. The views
    and conclusions contained in this document are those of the authors and
    should not be interpreted as representing the official policies, either
    expressed or implied, of the sponsoring organizations, agencies or the
    U.S.  government.}}

\author{
Hang Ma\\
Department of Computer Science\\
University of Southern California\\
hangma@usc.edu
\And
T.~K.~Satish Kumar\\
Department of Computer Science\\
University of Southern California\\
tkskwork@gmail.com
\And
Sven Koenig\\
Department of Computer Science\\
University of Southern California\\
skoenig@usc.edu
}

\maketitle

\begin{abstract}
  Several recently developed Multi-Agent Path Finding (MAPF) solvers scale to
  large MAPF instances by searching for MAPF plans on 2 levels: The high-level
  search resolves collisions between agents, and the low-level search plans
  paths for single agents under the constraints imposed by the high-level
  search.  We make the following contributions to solve the MAPF problem with
  imperfect plan execution with small average makespans: First, we formalize
  the MAPF Problem with Delay Probabilities (MAPF-DP), define valid MAPF-DP
  plans and propose the use of robust plan-execution policies for valid
  MAPF-DP plans to control how each agent proceeds along its path. Second, we
  discuss 2 classes of decentralized robust plan-execution policies (called
  Fully Synchronized Policies and Minimal Communication Policies) that prevent
  collisions during plan execution for valid MAPF-DP plans. Third, we present
  a 2-level MAPF-DP solver (called Approximate Minimization in Expectation)
  that generates valid MAPF-DP plans.
\end{abstract}

\section{Introduction}

Multi-Agent Path Finding (MAPF) is the problem of finding collision-free paths
for a given number of agents from their given start locations to their given
goal locations in a given environment. MAPF problems arise for aircraft towing
vehicles~\cite{airporttug16}, office
robots~\cite{DBLP:conf/ijcai/VelosoBCR15}, video game characters~\cite{WHCA}
and warehouse robots \cite{kiva}, among others.

Several recently developed MAPF solvers scale to large MAPF instances.
However, agents typically cannot execute their MAPF plans perfectly since
they often traverse their paths more slowly than intended. Their delay
probabilities can be estimated but current MAPF solvers do not use this
information, which often leads to frequent and runtime-intensive replanning or
plan-execution failures.

We thus formalize the MAPF Problem with Delay Probabilities (MAPF-DP), where
each agent traverses edges on an undirected graph (that models the
environment) to move from its start vertex to its goal vertex. At any discrete
time step, the agent can either execute 1) a {\em wait} action, resulting in
it staying in its current vertex, or 2) a {\em move} action with the intent of
traversing an outgoing edge of its current vertex, resulting in it staying in
its current vertex with the delay probability and traversing the edge
otherwise. The MAPF-DP problem is the problem of finding 1) a {\em MAPF-DP
  plan} that consists of a path for each agent from its start vertex to its
goal vertex (given by a sequence of wait and move actions) and 2) a {\em
  plan-execution policy} that controls with GO or STOP commands how each agent
proceeds along its path such that no collisions occur during plan
execution. There are 2 kinds of collisions, namely vertex collisions (where 2
agents occupy the same vertex at the same time step) and edge collisions
(where 2 agents traverse the same edge in opposite directions at the same time
step).

\begin{figure}[t]
  \centering
  \includegraphics[height=45pt]{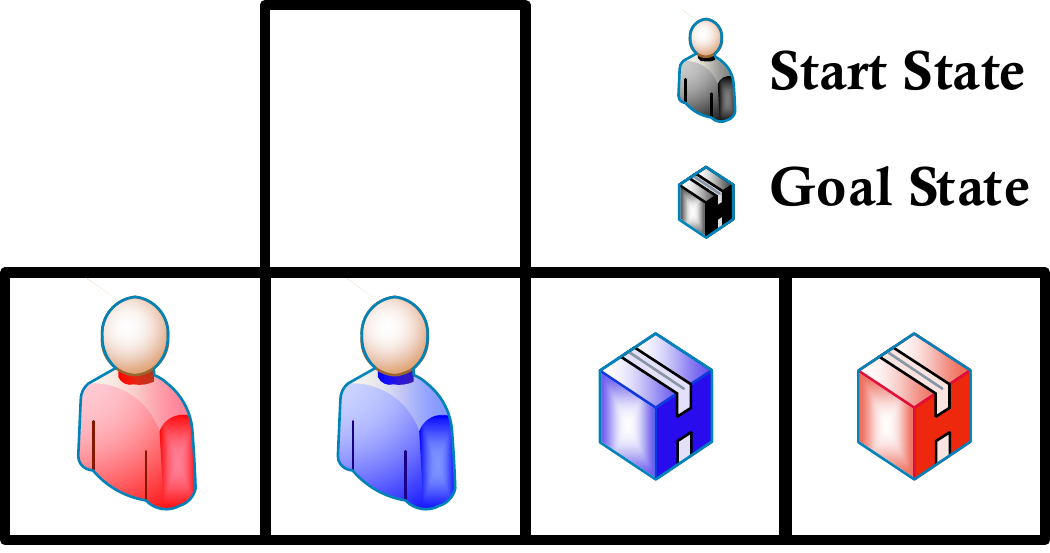} \hspace*{2mm}
  \raisebox{5pt}{\includegraphics[height=35pt]{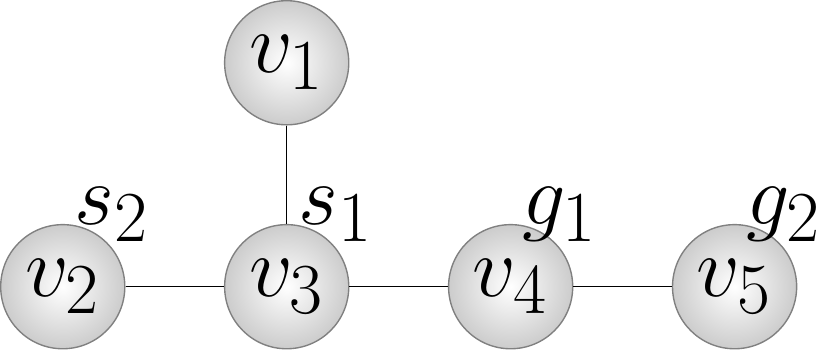}}\\
  \caption{A MAPF-DP instance.}\label{fig:MAPF}
\end{figure}

We make the following contributions to solve the MAPF-DP problem with small
average makespans: First, we formalize the MAPF-DP problem, define valid
MAPF-DP plans and propose the use of robust plan-execution policies for valid
MAPF-DP plans to control how each agent proceeds along its path. Second, we
discuss 2 classes of decentralized robust plan-execution policies (called
Fully Synchronized Policies and Minimal Communication Policies) that prevent
collisions during plan execution for valid MAPF-DP plans. Third, we present a
2-level MAPF-DP solver (called Approximate Minimization in Expectation) that
generates valid MAPF-DP plans.

\section{Background and Related Work}

The MAPF problem is NP-hard to solve optimally for flowtime minimization and
to approximate within any constant factor less than $4/3$ for makespan
minimization~\cite{MaAAAI16}. Search-based MAPF solvers can be optimal,
bounded suboptimal or
suboptimal~\cite{ODA,PushAndSwap,Wang11,EPEJAIR,DBLP:journals/ai/SharonSGF13,DBLP:journals/ai/SharonSFS15,ICBS,MStar,MaAAMAS16,CohenUK16}.
Current MAPF solvers typically assume perfect plan execution. However,
utilizing probabilistic information about imperfect plan execution can reduce
frequent and time-intensive replanning and plan-execution failures.

Partially Observable Markov Decision Processes (POMDPs) are a general
probabilistic planning framework. The MAPF-DP problem can be solved with
POMDPs but this is tractable only for very few agents in very small
environments since the size of the state space is proportional to the size of
the environment to the power of the number of agents and the size of the
belief space is proportional to the size of the state space to the power of
the length of the planning horizon \cite{SARSOP08,MaAAAI15}. Several
specialized probabilistic planning frameworks, such as transition-independent
decentralized Markov Decision Processes (Dec-MDPs)~\cite{Becker:2004} and
Multi-Agent Markov Decision Processes (MMDPs)~\cite{MMDP} can solve larger
probabilistic planning problems than POMDPs. In transition-independent
Dec-MDPs, the local state of each agent depends only on its previous local
state and the action taken by it \cite{Goldman04}. MAPF-DP is indeed
transition independent. However, there are interactions among agents since the
reward of each agent depends on whether it is involved in a collision and thus
on the local states of other agents and the actions taken by them. Fully
decentralized probabilistic planning frameworks thus cannot prevent
collisions. Fully centralized probabilistic planning frameworks can prevent
collisions but are more runtime-intensive and can thus scale poorly. For
example, the MAPF-DP problem can be solved with transition-independent
MMDPs~\cite{ScharpffROSW16}.  In fact, the most closely related research to
ours is that on approximating MMDPs \cite{LiuM16} although it handles
different types of dynamics than we do. The runtime of probabilistic planning
frameworks can be reduced by exploiting the problem structure, including when
interactions among agents are sparse. For example, decentralized
sparse-interaction Markov Decision Processes (Dec-SIMDPs) \cite{Melo11} assume
that interactions among agents occur only in well-defined interaction areas in
the environment (which is not the case for MAPF-DP in general), but typically
still do not scale to more than 10 agents. The model shaping technique for decentralized POMDPs \cite{VelagapudiVSS11} can compute policies for hundreds of agents greedily and UM* \cite{Wagner15} scales to
larger numbers of agents (with identical delay probabilities), but the plan
execution for both approaches is completely decentralized and thus cannot prevent collisions.

\section{Problem Definition: Planning}

A MAPF-DP instance is characterized by an undirected graph $G = (V, E)$ whose
vertices $V$ correspond to locations and whose edges $E$ correspond to
transitions between locations. We are given $m$ agents $a_1, a_2 \ldots
a_m$. Each agent $a_i$ has a unique start vertex $s_i \in V$, a unique goal
vertex $g_i \in V$ and a delay probability $p_i \in (0, 1)$.  A {\em path} for
agent $a_i$ is expressed by a function $l_i$ that maps each time index $x = 0,
1 \ldots X_i$ to a vertex $l_i(x) \in V$ such that $l_i(0) = s_i$, consecutive
vertices $l_i(x)$ and $l_i(x+1)$ are either identical (when agent $a_i$ is
scheduled to execute a wait action) or connected by an edge (when agent $a_i$
is scheduled to execute a move action from vertex $l_i(x)$ to vertex
$l_i(x+1)$) and $l_i(X_i) = g_i$. A {\em MAPF plan} consists of a path for
each agent.

\section{Problem Definition: Plan Execution}

The local state $x_i^t$ of agent $i$ at time step $t = 0, 1 \ldots \infty$
during plan execution is a time index. We set $x_i^0 := 0$ and always update
its local state such that it is in vertex $l_i(x_i^t)$ at time step $t$. The
agent knows its current local state and receives messages from some of the
other agents about their local states. At each time step, its plan-execution
policy maps this knowledge to one of the commands $GO$ or $STOP$ that control
how it proceeds along its path.

\begin{enumerate}[nolistsep]
\item If the command is $GO$ at time step $t$:
\begin{enumerate}[nolistsep]
\item If $x_i^t = X_i$, then agent $a_i$ executes no action and remains in its
  current vertex $l_i(x_i^t)$ since it has entered its last local state (and
  thus the end of its path). We thus update its local state to $x_i^{t+1} :=
  x_i^t$.
\item If $x_i^t \neq X_i$ and $l_i(x_i^t) = l_i(x_i^t + 1)$, then agent $a_i$
  executes a wait action to remain in its current vertex $l_i(x_i^t)$. The
  execution of wait actions never fails. We thus update its local state to
  $x_i^{t+1} := x_i^t + 1$ (success).
\item If $x_i^t \neq X_i$ and $l_i(x_i^t) \neq l_i(x_i^t + 1)$, then agent
  $a_i$ executes a move action from its current vertex $l_i(x_i^t)$ to vertex
  $l_i(x_i^t + 1)$. The execution of move actions fails with delay probability
  $p_i$ with the effect that the agent executes no action and remains delayed
  in its current vertex $l_i(x_i^t)$. We thus update its local state to
  $x_i^{t+1} := x_i^t$ with probability $p_i$ (failure) and $x_i^{t+1} :=
  x_i^t + 1$ with probability $1-p_i$ (success).
\end{enumerate}
\item If the command is $STOP$ at time step $t$, then agent $a_i$ executes no
  action and remains in its current vertex $l_i(x_i^t)$. We thus update its
  local state to $x_i^{t+1} := x_i^t$.
\end{enumerate}

Our objective is to find a combination of a MAPF plan and a plan-execution
policy with small average makespan, which is the average earliest time step
during plan execution when all agents have entered their last local
states. The MAPF problem is a special case where the delay probabilities of
all agents are zero and the plan-execution policies always provide GO
commands.

\section{Valid MAPF-DP Plans}

\begin{defn}\label{valid_plan}
A {\em valid MAPF-DP plan} is a plan with 2 properties:
\begin{enumerate}[nolistsep]
\item $\forall i, j, x~\mbox{with}~i \neq j:~l_i(x)\neq l_j(x)$ [two agents
  are never scheduled to be in the same vertex at the same time index, that
  is, the vertices of two agents in the same local state are different].
\item $\forall i, j, x~\mbox{with}~i \neq j:~l_i(x+1) \neq l_j(x)$ [an agent
  is never scheduled to be in a vertex at a time index $x+1$ when any other
  agent is scheduled to be in the same vertex at time index $x$, that is, the
  vertex of an agent in a local state $x+1$ has to be different from the
  vertex of any other agent in local state $x$].
\end{enumerate}
\end{defn}

Figure~\ref{fig:MAPF} shows a sample MAPF-DP instance where the blue agent
$a_1$ has to move from its start vertex $v_3$ to its goal vertex $v_4$ and the
red agent $a_2$ has to move from its start vertex $v_2$ to its goal vertex
$v_5$. Agent $a_1$ has to move north to let agent $a_2$ pass. The paths $l_1 =
\langle v_3, v_1, v_1, v_1, v_3, v_4 \rangle$ and $l_2 = \langle v_2, v_2,
v_3, v_4, v_5 \rangle$ form a valid MAPF-DP plan. However, the paths $l_1$ $=$
$\langle v_3, v_1, v_1, v_3, v_4 \rangle$ and $l_2$ $=$ $\langle v_2, v_3, v_4, v_5
\rangle$ a valid MAPF plan but not a valid MAPF-DP plan since $l_2(1) = l_1(0)
= v_3$ violates Property 2.

Property 1 of Definition \ref{valid_plan} is necessary to be able to execute
valid MAPF-DP plans without vertex collisions because two agents could
otherwise be in the same vertex at the same time step (under perfect or
imperfect plan execution). Property 2 is also necessary because an agent could
otherwise enter the vertex of some other agent that unsuccessfully tries to
leave the same vertex at the same time step (under imperfect plan execution).
Property 2 is also necessary to be able to execute valid MAPF-DP plans without
edge collisions (under perfect or imperfect plan execution).

\section{Robust Plan-Execution Policies}

We study 2 kinds of decentralized {\em robust plan-execution policies} for
valid MAPF-DP plans, which are plan-execution policies that prevent all
collisions during the imperfect plan execution of valid MAPF-DP plans.

\subsection{Fully Synchronized Policies (FSPs)}

Fully Synchronized Policies (FSPs) attempt to keep all agents in lockstep as
much as possible by providing a GO command to an agent if and only if the
agent has not yet entered its last local state and all other agents have
either entered their last local states or have left all local states that
precede the local state of the agent itself. FSPs can be implemented easily if
each agent sends a message to all other agents when it enters a new local
state. An agent can implement its FSP simply by counting how many messages it
has received from each other agent and providing a GO command to itself in
local state $x$ if and only if it has not yet entered its last local state and
has received $x$ messages over the course of plan execution from each other
agent.

\subsection{Minimal Communication Policies (MCPs)}

FSPs have 2 drawbacks. First, agents wait unnecessarily, which results in
large average makespans. Second, each agent always needs to know the local
states of all other agents, which results in many sent messages. Property 2 of
Definition \ref{valid_plan} suggests that robust plan-execution policies for
valid MAPF-DP plans could provide a GO command to an agent if and only if the
agent has not yet entered its last local state and all other agents have left
all local states that precede the local state of the agent itself and whose
vertices are the same as the vertex of the next local state of the agent
itself. This way, it is guaranteed that the vertex of the next local state of
the agent is different from the vertices of all other agents in their current
local states. Minimal Communication Policies (MCPs) address these drawbacks by
identifying such critical dependencies between agents and obeying them during
plan execution, an idea that originated in the context of centralized
non-robust plan-execution policies \cite{HoenigICAPS16}.

The local state of an agent $a_i$ at any time step during plan execution is a
time index $x$. Since we need to relate the local states of different agents,
we use $l_i(x)$ in the following not only to refer to the vertex assigned to
local state $x$ of agent $a_i$ but also to the local state $x$ of agent $a_i$
itself (instead of $x$), depending on the context.

Every valid MAPF-DP plan defines a total order on the local states of all
agents, which we relax to a partial order $\rightarrow$ as follows:

\begin{enumerate}[nolistsep]
\item $\forall i, x: l_i(x) \rightarrow l_i(x+1)$ [agent $a_i$ enters a local
  state $x+1$ during plan execution only after it enters local state $x$].
\item $\forall i, j, x, x'$ with $i \neq j$, $x' < x$ and $l = l_j(x') =
  l_i(x+1) : l_j(x'+1) \rightarrow l_i(x+1)$ [agent $a_i$ enters a local state
  $x+1$ with a vertex $l$ during plan execution only after agent $a_j$ has
  left a local state $x'$ with vertex $l$ (and thus entered local state
  $x'+1$) that precedes local state $x$].
\end{enumerate}

Property 1 of the partial order enforces that each agent visits its locations in the same order as in the MAPF-DP plan. Property 2 enforces that any two agents visit the same location in the same order as in the MAPF-DP plan. We can express the partial order with a directed graph $\mathcal{G} = (\mathcal{V}, \mathcal{E})$ whose vertices correspond to local states and
whose edges correspond to the partial order given by the two properties
above. Property 2 specifies the critical dependencies between agents. Edges
are redundant and can then be removed from the directed graph when they are
implied by the other edges due to transitivity. A {\em transitive reduction}
of the directed graph minimizes the number of remaining edges. It can be
computed in time $O(|\mathcal{V}||\mathcal{E}|)$ \cite{AhoGU72}, is unique,
contains all edges between local states of the same agent (since they are
never redundant) and thus minimizes the number of edges between the local
states of different agents.

MCPs can be implemented easily if each agent $a_j$ sends a message to each
other agent $a_i$ when agent $a_j$ enters a new local state $\bar{x}'$ (=
$x'+1$ in Property 2) if and only if the transitive reduction contains an edge
$l_j(\bar{x}') \rightarrow l_i(\bar{x})$ for some local state $\bar{x}$ (=
$x+1$ in Property 2) of agent $a_i$. Since the transitive reduction minimizes
the number of edges between the local states of different agents, it also
minimizes the number of sent messages. An agent $a_i$ can implement its MCP
simply by counting how many messages it has received from each other agent and
providing a GO command to itself in local state $x$ if and only if it has not
yet entered its last local state and has received a number of messages over
the course of plan execution from each other agent $a_j$ that corresponds to
the number of incoming edges from local states of agent $a_j$ to its local
states $0 , 1 \ldots x+1$.

\begin{figure}[t]
  \centering
  \includegraphics[width=0.9\columnwidth]{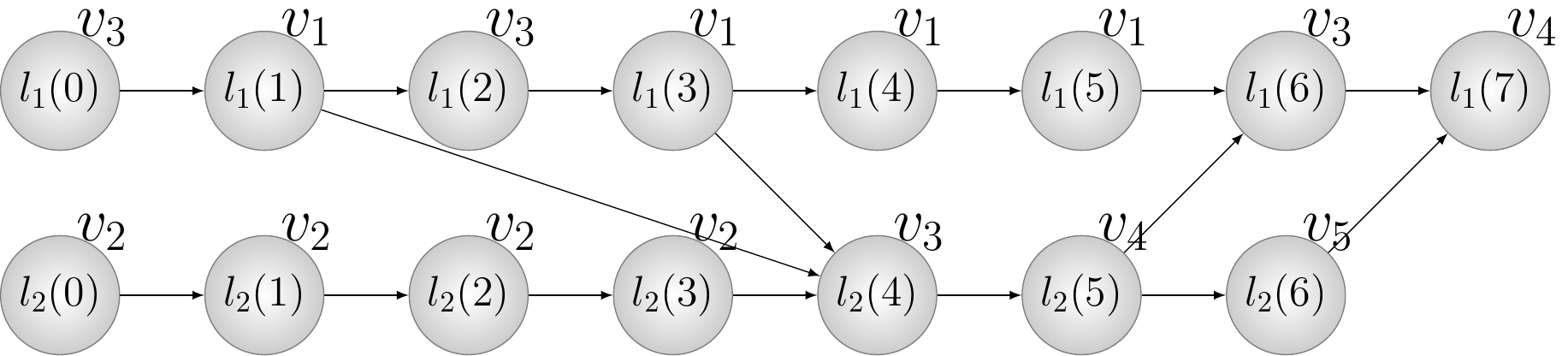}\\
  \caption{A directed graph that specifies a partial order on the local states
    for the MAPF-DP instance from Figure~\ref{fig:MAPF} and its valid MAPF-DP
    plan $l_1 = \langle v_3, v_1, v_3, v_1, v_1, v_1, v_3, v_4 \rangle$ and
    $l_2 = \langle v_2, v_2, v_2, v_2, v_3, v_4, v_5
    \rangle$.} \label{fig:dgraph1}
\end{figure}

\begin{figure}[t]
  \centering
  \includegraphics[width=0.9\columnwidth]{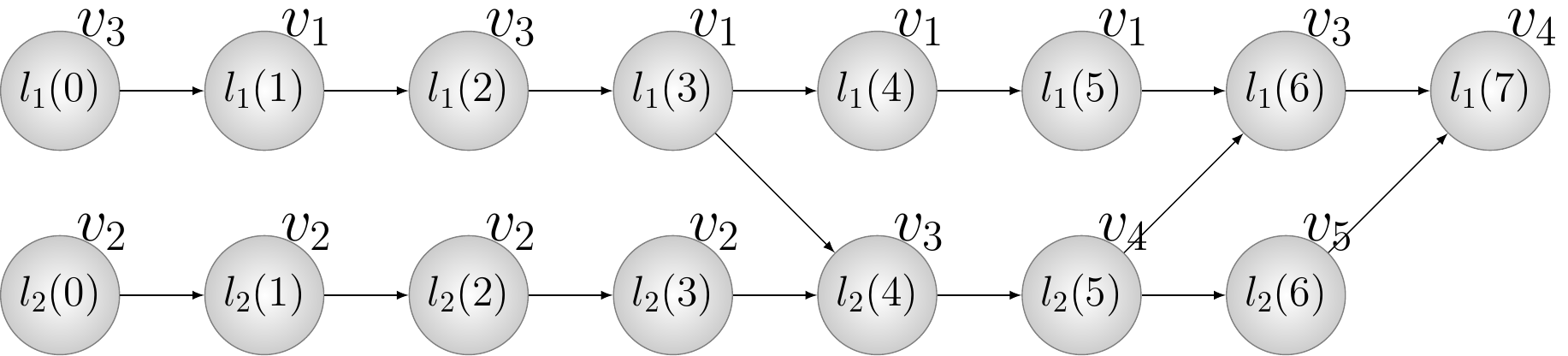}
  \caption{The transitive reduction for Figure~\ref{fig:dgraph1}.}\label{fig:dgraph2}
\end{figure}

Figure \ref{fig:dgraph1} shows a sample partial order on the local states for
the MAPF-DP instance from Figure \ref{fig:MAPF} and its valid MAPF-DP plan
$l_1 = \langle v_3, v_1, v_3, v_1, v_1, v_1, v_3, v_4 \rangle$ and $l_2 =
\langle v_2, v_2, v_2, v_2, v_3, v_4, v_5 \rangle$.  $l_1(1) \rightarrow
l_2(4)$, for example, is implied by $l_1(1) \rightarrow l_1(2) \rightarrow
l_1(3) \rightarrow l_2(4)$ and can thus be removed from the directed graph.
Figure~\ref{fig:dgraph2} shows the resulting transitive reduction, which
implies that agent $a_2$ has to wait in local state $3$ until it has received
one message from agent $a_1$ during the course of plan execution but can then
proceed through all future local states without waiting.

\subsection{Properties of FSPs and MCPs}

Both FSPs and MCPs do not result in deadlocks during the plan execution of
valid MAPF-DP plans because there always exists at least one agent that is
provided a GO command before all agents have entered their last local states
(namely an agent with the smallest local state among all agents that have not
yet entered their last local states since an agent can wait only for other
agents with smaller local states).

Both FSPs and MCPs are robust plan-execution policies due to Properties 1 and
2 of valid MAPF-DP plans. We now provide a proof sketch for the robustness of
MCPs.

First, consider a valid MAPF-DF plan and assume that $l_i(x) = l_j(y)$ for two
agents $a_i$ and $a_j$ with $i \neq j$. Then, 1) $y \neq x$ since $l_i(x) \neq
l_j(x)$ according to Property 1 of Definition \ref{valid_plan} and 2) $y \neq
x+1$ since $l_j(x+1) \neq l_i(x)$ according to Property 2 of Definition
\ref{valid_plan} (State Property).

Second, we show by contradiction that no vertex collisions can occur during
plan execution. Assume that a vertex collision occurs between agents $a_i$ and
$a_j$ with $i \neq j$ when agent $a_i$ is in local state $x$ and agent $a_j$
is in local state $y$. Assume without loss of generality that $x \leq
y$. Then, $l_i(x+1) \rightarrow l_j(y)$ according to Property 2 of the partial
order $\rightarrow$ since $l_i(x) = l_j(y)$ according to our vertex collision
assumption and $x < y-1$ according to the State Property. Thus, agent $a_j$ can
leave local state $y-1$ only when agent $a_i$ reaches local state $x+1$, which
is a contradiction with the vertex collision assumption.

Third, we show by contradiction that no edge collisions can occur during plan
execution. Assume that an edge collision between agents $a_i$ and $a_j$ with
$i \neq j$ occurs when agent $a_i$ changes its local state from $x$ to $x+1$
and agent $a_j$ changes its local state from $y$ to $y+1$. Assume without loss
of generality that $x \leq y$. Case 1) If $x = y$, then $l_i(x) = l_j(y+1) =
l_j(x+1)$, which is a contradiction with the State Property. Case 2) If $x <
y$, then $l_i(x+1) \rightarrow l_j(y+1)$ according to Property 2 of the
partial order $\rightarrow$ since $l_i(x) = l_j(y+1)$ according to our edge
collision assumption and $x < y$ according to the case assumption. Thus, agent
$a_j$ can leave local state $y$ only when agent $a_i$ reaches local state
$x+1$, which is a contradiction with the edge collision assumption.

\section{Approximate Minimization in Expectation}

MCPs are robust plan-execution policies for valid MAPF-DP plans that do not
stop agents unnecessarily and result in few sent messages. We present a
MAPF-DP solver, called Approximate Minimization in Expectation (AME), that
determines valid MAPF-DP plans so that their combination with MCPs results in
small average makespans.

AME is a 2-level MAPF-DP solver that is based on Conflict-Based Search (CBS)
\cite{DBLP:journals/ai/SharonSFS15}. Its high-level search imposes constraints
on the low-level search that resolve violations of Properties 1 and 2 of
Definition~\ref{valid_plan} (called {\em conflicts}). Its low-level search
plans paths for single agents that obey these constraints and result in small
average makespans. The average makespan of a MAPF-DP plan is the expectation
of the maximum of (one or more) random variables that represent the time steps
when all agents enter their last local states. Moreover, the average time step
when an agent enters a local state is the expectation of the maximum of random
variables as well. It is often difficult to obtain good closed-form
approximations of the expectation of the maximum of random variables. AME thus
approximates it with the maximum over the expectations of the random
variables, which typically results in an underestimate but, according to our
experimental results, a close approximation. The approximate average time step
$\tilde{l}_i(x)$ when agent $a_i$ enters a local state $x$ for a given MAPF-DP
plan is 0 for $x=0$ and { \footnotesize

\begin{eqnarray}
  & \max(\tilde{l}_i(x-1), \max_{j,x':i \neq j,x'<x,l_j(x') \rightarrow
    l_i(x)}(\tilde{l}_j(x'))) + \hat{t}_i \nonumber \\
  & = \max_{j,x':x'<x,l_j(x') \rightarrow l_i(x)}(\tilde{l}_j(x')) + \hat{t}_i
  \label{eq1}
\end{eqnarray}}

\noindent otherwise since agent $a_i$ first enters local state $x-1$ at
approximate average time step $\tilde{l}_i(x-1)$, then might have to wait for
messages from other agents $a_j$ that they send when they enter their local
states $x'$ at approximate average time steps $\tilde{l}_j(x')$ and finally
has to successfully execute one action (perhaps repeatedly) to enter local
state $x$. The average number $\hat{t}_i$ of time steps that it needs for the
successful execution of the action is 1 (for a wait action) if $l_i(x) =
l_i(x-1)$ and $1/(1-p_i)$ (for a move action) otherwise. The approximate
average makespan of the given MAPF-DP plan is then $\max_i \tilde{l}_i(X_i)$
since all agents need to enter their last local states. One might be able to
obtain better approximations with more runtime-intensive importance sampling
or dynamic programming methods but the runtime of the resulting AME variant
would be large since it needs to compute many such approximations.

\begin{algorithm}[t]
\tiny
\renewcommand\arraystretch{0.4}
\caption{High-Level Search of AME.}
\label{High-Level Search}
    $\mbox{\it Root.constraints} := \emptyset$\;
    $\mbox{\it Root.plan} := \emptyset$\;
    \For{\textnormal{\textbf{each} agent $a_i$}}
    {
      \If{\textnormal{LowLevelSearch($a_i$, $\mbox{\it Root}$, 0) returns no path (nor its labels)}}
      {
        \Return ``No solution exists''\;
      }
      Add the returned path (and its labels) to $\mbox{\it Root.plan}$\;
    }
    $\mbox{\it Root.key} :=$ ApproximateAverageMakespan({\it Root.plan})\;
    $\mbox{\it Priorityqueue} := \{\mbox{\it Root}\}$\;
    \While{$\mbox{\it Priorityqueue} \neq \emptyset$}
    {
        $N := \mbox{\it Priorityqueue}$.pop()\;
        \If{\textnormal{FindConflicts($N.\mbox{\it plan}$) returns no conflicts}}
        {
            \Return ``Solution is'' $N.\mbox{\it plan}$\;
        }
        $\mbox{\it Conflict} :=$ earliest returned conflict\;
        \For{\textnormal{\textbf{each} agent $a_i$ involved in $\mbox{\it Conflict}$}}
        {
            $N' :=$ new node with parent node $N$\;
            $N'.\mbox{\it constraints} := N.\mbox{\it constraints}$\;
            $N'.\mbox{\it plan} := N.\mbox{\it plan}$\;
            Add one new constraint for agent $a_i$ to $N'.\mbox{\it constraints}$ (see main text)\;
            \If{\textnormal{LowLevelSearch($a_i$, $N'$, $N.\mbox{\it key}$) returns a path (and its labels)}}
            {
                Replace the path (and its labels) of agent $a_i$ in
                $N'.\mbox{\it plan}$ with the returned path (and its labels)\;
                $N'.\mbox{\it key} :=$ ApproximateAverageMakespan($N'.\mbox{\it plan}$)\;
                $\mbox{\it Priorityqueue}$.insert($N'$)\;
            }
        }
    }
    \Return ``No solution exists''\;
\end{algorithm}

\subsection{High-Level Search}

Algorithm~\ref{High-Level Search} shows the high-level search of AME, which is
similar to the high-level search of CBS. In the following, we point out the
differences. Each high-level node $N$ contains the following items:
\begin{enumerate}[nolistsep]
\item A set $N.\mbox{\it constraints}$ of constraints of the form $(a_i, l, x)$ that
  states that the vertex of agent $a_i$ in local state $x$ has to be different
  from vertex $l$.
\item A (labeled) MAPF-DP plan $N.\mbox{\it plan}$ that contains a path $l_i$ for each
  agent $a_i$ (that obeys the constraints $N.\mbox{\it constraints}$) and an
  approximation $\tilde{l}_i(x)$ (called label) of each average time step when
  agent $a_i$ enters local state $x$ during plan execution with MCPs.
\item The key $N.\mbox{\it key}$ of high-level node $N$ that encodes its priority
  (smaller keys have higher priority) and is equal to the approximate average
  makespan of MAPF-DP plan $N.\mbox{\it plan}$ given by
  ApproximateAverageMakespan$(N.\mbox{\it plan})$ $=$ $\max_{i} \tilde{l}_i(X_i)$.
\end{enumerate}

When a conflict exists in MAPF-DP plan $N.\mbox{\it plan}$, then the high-level search
creates 2 child nodes of node $N$ [Line 15] whose constraints are initially
set to the constraints $N.\mbox{\it constraints}$ [Line 16] and whose MAPF-DP plan is
initially set to MAPF-DP plan $N.\mbox{\it plan}$ [Line 17]. Assume that the earliest
conflict is a violation of Property 1 in Definition~\ref{valid_plan}, in which
case the vertices of two agents $a_i$ and $a_j$ in a local state $x$ are both
identical to a vertex $l$. In this case, AME adds the constraint $(a_i, l, x)$
to the constraints of the first child node and the constraint $(a_j, l, x)$ to
the constraints of the second child node [Line 18], thus preventing the
conflict in both cases. Assume that the earliest conflict is a violation of
Property 2 in Definition~\ref{valid_plan}, in which case the vertex of an
agent $a_i$ in a local state $x+1$ and the vertex of some other agent $a_j$ in
the immediately preceding local state $x$ are both identical to a vertex $l$.
In this case, AME adds the constraint $(a_i, l, x+1)$ to the constraints of
the first child node and the constraint $(a_j, l, x)$ to the constraints of
the second child node [Line 18], thus preventing the conflict in both cases.

\subsection{Low-Level Search}

LowLevelSearch($a_i$, $N$, $\mbox{\it key}$) finds a new path for agent $a_i$ and the
labels $\tilde{l}_i(x)$ of this path. It uses the paths of the other agents
and their labels in $N.\mbox{\it plan}$ but does not update them. (The paths are empty
directly after the execution of Line 2.) It performs a focal search with
re-expansions in a state space whose states correspond to pairs of vertices
and local states (except for those pairs ruled out by constraints in
$N.\mbox{\it constraints}$ that pertain to agent $a_i$) and whose edges connect state
$(l, x)$ to state $(l', x+1)$ if and only if $l = l'$ (for a wait action) or
$(l,l') \in E$ (for a move action). The g-value of a state $(l,x)$
approximates (sic!) the approximate average time step $\tilde{l}_i(x)$. The
start state is $(s_i,0)$ and its g-value is 0. When the low-level search
expands state $(l,x-1)$, it sets the g-value of its successor $(l',x)$
according to Equation (\ref{eq1}) to the minimum of its current g-value
$g((l',x))$ and

{\footnotesize
\begin{eqnarray*}
\max(g((l,x-1)), max_{j,x':i \neq j,x'<x,l_j(x') \rightarrow l_i(x)}(\tilde{l}_j(x'))) + \hat{t}_i,
\end{eqnarray*}}

\noindent where $\hat{t}_i$ is 1 if $l = l'$ and $1/(1-p_i)$ otherwise. The low-level
search decides which state $(l,x)$ to expand next based on 1) the f-value of
the state, which is the sum of its g-value and its h-value, where the h-value
is $1/(1-p_i)$ times the distance from location $l$ to location $g_i$ in graph
$G$ (which is an optimistic estimate of the average number of time steps
required to move from location $l$ to location $g_i$) and 2) the number of
conflicts of the path for agent $a_i$ that corresponds to the locations in the
states on the found path from the start state to $(l,x)$ with the paths of
other agents.

The low-level search starts in Phase 1. The objective in this phase is to find
a path for agent $a_i$ so that it enters its last local state with a
reasonably small approximate average number of time steps, namely one that is
no larger than the approximate average makespan $\mbox{\it key}$ of the MAPF-DP plan in
the parent node of node $N$ in the high-level search, and has a small number
of conflicts. The first part of the objective tries to ensure that the
approximate average makespan of the resulting MAPF-DP plan in node $N$ is no
larger than the one of the MAPF-DP plan in the parent node of node $N$, and
the second part tries to ensure that the resulting MAPF-DP plan has a small
number of conflicts so that the high-level search has a small runtime since it
needs to resolve only a small number of conflicts.  The low-level search thus
repeatedly expands a state with the smallest number of conflicts among all
states in the priority queue whose f-values are no larger than $\mbox{\it key}$.

If no such state exists, then the low-level search switches to Phase 2. The
objective in this phase is to find a path for agent $a_i$ so that it enters
its last local state with a small approximate average number of time
steps. This objective tries to ensure that the approximate average makespan of
the resulting MAPF-DP plan in node $N$ is not much larger than the one of the
MAPF-DP plan in the parent node of node $N$. The low-level search thus
repeatedly expands a state with the smallest f-value among all states in the
priority queue.

The low-level search terminates successfully when it is about to expand a
state $(l,x)$ with $l = g_i$ and $N.\mbox{\it constraints}$ contains no constraints of
the form $(a_i,g_i,x')$ with $x' > x$. It then sets $X_i := x$, the locations
$l_i(x)$ that form the path of agent $a_i$ to the corresponding locations in
the states on the found path from the start state to $(l,x)$ and the
approximate average time steps $\tilde{l}_i(x)$ to the corresponding g-values
of these states. The low-level search terminates unsuccessfully when the
priority queue becomes empty. The low-level search currently does not
terminate otherwise but we might be able to make it complete by using an upper
bound on the smallest average makespan of any valid MAPF-DP plan, similar to
upper bounds in the context of valid MAPF plans \cite{Kornhauser1984}.

\subsection{Future Work}

The low-level search is currently the weakest part of AME due to the many
approximations to keep its runtime small which is important since the
high-level search runs many low-level searches. We expect that future work
will be able to improve the low-level search substantially. For example, the
approximate average time steps $\tilde{l}_j(x)$ for agents $a_j$ different from agent
$a_i$ could be updated before, during or after the local search, which would
provide more accurate values for the current and future low-level searches as
well as the current high-level search. Once the low-level search finds a path
for agent $a_i$ and the high-level search replaces the path of agent $a_i$ in
the MAPF-DP plan in the current high-level node with this path, it could
update the approximate average time steps of all agents to the ideal
approximate average time steps given by Equations (\ref{eq1}), for example as
part of the execution of ApproximateAverageMakespan on Lines 7 and 21. Many
other improvements are possible as well.

\section{Experiments}

We evaluate AME with MCPs on a 2.50 GHz Intel Core i5-2450M PC with 6 GB RAM.

\begin{figure}[t]
  \centering
  \includegraphics[width=0.5\columnwidth]{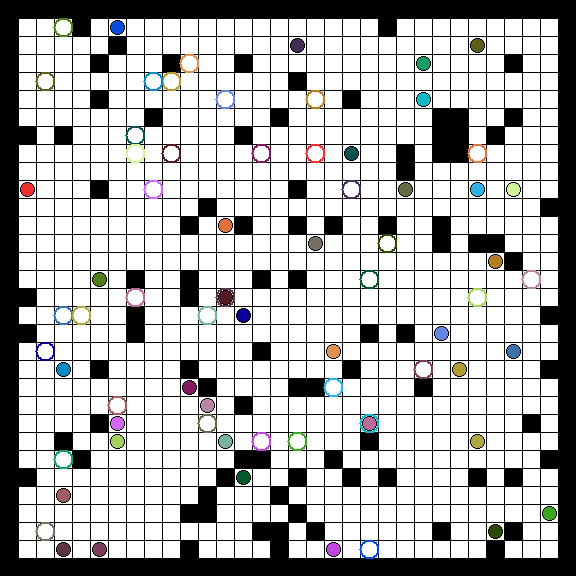}\\
  \includegraphics[width=\columnwidth]{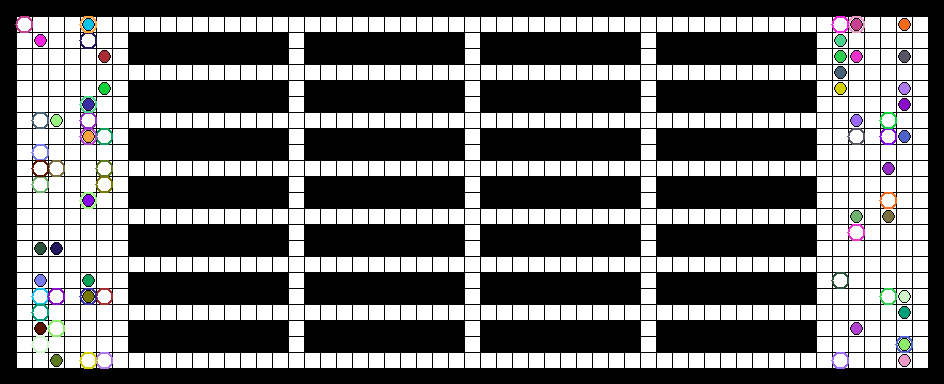}
  \caption{Two MAPF-DP instances: random 1 (top) and warehouse 1 (bottom). Blocked cells are shown in black. The start and goal cells for each agent are represented by a solid circle and a hollow circle of the same color, respectively.}\label{fig:instance}
\end{figure}

\subsection{Experiment 1: MAPF Solvers}

\begin{table}[t]
\centering
\caption{Results of different MAPF(-DP) solvers for MAPF-DP instances with 35
  agents and delay probability range $(0, 1/2)$.}
\label{tab:1}
\fontsize{50}{60}\selectfont
\resizebox{\columnwidth}{!}{
\begin{tabular}{c|cccc|ccc|ccc}
\hline
\multicolumn{1}{>{\hspace{-5pt}}c<{\hspace{-5pt}}|}{}   & \multicolumn{4}{>{\hspace{-5pt}}c<{\hspace{-5pt}}|}{AME}                                                                                                                                                                                                             & \multicolumn{3}{>{\hspace{-5pt}}c<{\hspace{-5pt}}|}{Push and Swap}                                                                                                        & \multicolumn{3}{>{\hspace{-5pt}}c<{\hspace{-5pt}}}{Adapted CBS}                                                                                                          \\
\hline
\multicolumn{1}{>{\hspace{-5pt}}c<{\hspace{-5pt}}|}{id} &
\multicolumn{1}{>{\hspace{-5pt}}c<{\hspace{-5pt}}}{\begin{tabular}[c]{>{\hspace{-5pt}}c<{\hspace{-5pt}}}runtime\\ (s)\end{tabular}} &
\multicolumn{1}{>{\hspace{-5pt}}c<{\hspace{-5pt}}}{\begin{tabular}[c]{>{\hspace{-5pt}}c<{\hspace{-5pt}}}approx-\\imate\\average\\makespan\end{tabular}} &
\multicolumn{1}{>{\hspace{-5pt}}c<{\hspace{-5pt}}}{\begin{tabular}[c]{>{\hspace{-5pt}}c<{\hspace{-5pt}}}average\\makespan\end{tabular}} &
\multicolumn{1}{>{\hspace{-5pt}}c<{\hspace{-5pt}}|}{\begin{tabular}[c]{>{\hspace{-5pt}}c<{\hspace{-5pt}}}mess-\\ages\end{tabular}} &
\multicolumn{1}{>{\hspace{-5pt}}c<{\hspace{-5pt}}}{\begin{tabular}[c]{>{\hspace{-5pt}}c<{\hspace{-5pt}}}runtime\\(s)\end{tabular}} &
\multicolumn{1}{>{\hspace{-5pt}}c<{\hspace{-5pt}}}{\begin{tabular}[c]{>{\hspace{-5pt}}c<{\hspace{-5pt}}}average\\makespan\end{tabular}} &
\multicolumn{1}{>{\hspace{-5pt}}c<{\hspace{-5pt}}|}{\begin{tabular}[c]{>{\hspace{-5pt}}c<{\hspace{-5pt}}}mess-\\ages\end{tabular}} &
\multicolumn{1}{>{\hspace{-5pt}}c<{\hspace{-5pt}}}{\begin{tabular}[c]{>{\hspace{-5pt}}c<{\hspace{-5pt}}}runtime\\(s)\end{tabular}} & \multicolumn{1}{>{\hspace{-5pt}}c<{\hspace{-5pt}}}{\begin{tabular}[c]{>{\hspace{-5pt}}c<{\hspace{-5pt}}}average\\ makespan\end{tabular}} & \multicolumn{1}{>{\hspace{-5pt}}c<{\hspace{-5pt}}}{\begin{tabular}[c]{>{\hspace{-5pt}}c<{\hspace{-5pt}}}mess-\\ages\end{tabular}} \\
\hline
random 1  & 0.058 & 63.15 & 71.28 $\pm$ 0.34 & 267 & 0.031 & 812.41 $\pm$ 0.40 & 287 & -       & -                & -   \\
random 2  & 0.052 & 66.22 & 73.02 $\pm$ 0.29 & 257 & 0.025 & 768.30 $\pm$ 0.43 & 257 & -       & -                & -   \\
random 3  & 0.080  & 78.44 & 84.90 $\pm$ 0.40 & 373 & 0.052 & 934.59 $\pm$ 0.33 & 387 & -       & -                & -   \\
random 4  & 0.063 & 67.00 & 72.89 $\pm$ 0.37 & 251 & 0.028 & 755.95 $\pm$ 0.33 & 255 & -       & -                & -   \\
random 5  & 0.050  & 65.13 & 73.98 $\pm$ 0.31 & 255 & 0.029 & 875.48 $\pm$ 0.47 & 318 & 282.079 & 84.11 $\pm$ 0.40 & 282 \\
random 6  & 0.052 & 62.89 & 66.98 $\pm$ 0.36 & 257 & 0.031 & 830.77 $\pm$ 0.32 & 290 & -       & -                & -   \\
random 7  & 0.495 & 67.22 & 71.34 $\pm$ 0.36 & 269 & 0.038 & 785.55 $\pm$ 0.46 & 274 & -       & -                & -   \\
random 8  & 0.042 & 49.33 & 51.72 $\pm$ 0.35 & 164 & 0.024 & 648.80 $\pm$ 0.35 & 199 & 197.911 & 52.35 $\pm$ 0.37 & 163 \\
random 9  & 0.051 & 56.27 & 61.30 $\pm$ 0.27 & 247 & 0.052 & 780.60 $\pm$ 0.30 & 294 & -       & -                & -   \\
random 10 & 0.487 & 60.06 & 64.77 $\pm$ 0.38 & 234 & 0.032 & 750.12 $\pm$ 0.35 & 284 & -       & -                & -   \\
\hline
warehouse 1  & 0.124 & 114.32 & 124.18 $\pm$ 0.44 & 705 & 0.055 & 1,399.14 $\pm$ 0.43 & 703 & - & - & - \\
warehouse 2  & 0.106 & 119.74 & 124.63 $\pm$ 0.51 & 762 & 0.055 & 1,620.03 $\pm$ 0.60 & 810 & - & - & - \\
warehouse 3  & 0.107 & 112.96 & 117.00 $\pm$ 0.53 & 609 & 0.032 & 1,295.75 $\pm$ 0.53 & 616 & - & - & - \\
warehouse 4  & 0.090  & 114.90 & 117.31 $\pm$ 0.52 & 541 & 0.043 & 1,246.47 $\pm$ 0.67 & 571 & - & - & - \\
warehouse 5  & -     & -      & -                 & -   & 0.060  & 1,453.36 $\pm$ 0.54 & 783 & - & - & - \\
warehouse 6  & 0.111 & 127.65 & 131.10 $\pm$ 0.59 & 710 & 0.037 & 1,437.01 $\pm$ 0.58 & 664 & - & - & - \\
warehouse 7  & 0.142 & 87.45  & 96.54 $\pm$ 0.34  & 488 & 0.028 & 1,154.21 $\pm$ 0.60 & 403 & - & - & - \\
warehouse 8  & -     & -      & -                 & -   & 0.024 & 1,233.13 $\pm$ 0.58 & 401 & - & - & - \\
warehouse 9  & 0.087 & 103.51 & 107.33 $\pm$ 0.42 & 462 & 0.024 & 1,088.53 $\pm$ 0.44 & 422 & - & - & - \\
warehouse 10 & 0.183 & 120.76 & 127.36 $\pm$ 0.53 & 909 & 0.057 & 1,541.56 $\pm$ 0.62 & 678 & - & - & - \\
\hline
\end{tabular}
}
\end{table}

We compare AME to 2 MAPF solvers, namely 1) Adapted CBS, a CBS variant that
assumes perfect plan execution and computes valid MAPF-DP plans, minimizes $\max_i X_i$ and breaks ties toward
paths with smaller $X_i$ and thus fewer actions and 2) Push and
Swap~\cite{PushAndSwap}, a MAPF solver that assumes perfect plan execution and
computes valid MAPF-DP plans where exactly one agent executes a move action at
each time step and all other agents execute wait actions. We generate 10
MAPF-DP instances (labeled random 1-10) in 30$\times$30 4-neighbor grids with 10\% randomly blocked
cells and random but unique start and unique goal cells for 35 agents whose delay probabilities for AME are sampled uniformly at random from the delay
probability range $(0, 1/2)$. In the same way, we generate 10 MAPF-DP instances (labeled warehouse 1-10) in a simulated warehouse environment with random but unique start and unique goal cells on the left and right sides. Figure~\ref{fig:instance} shows two MAPF-DP instances: random 1 (top) and warehouse 1 (bottom).

Table~\ref{tab:1} reports for each MAPF-DP instance the runtime, the
approximate average makespan calculated by AME, the average makespan over
1,000 plan-execution runs with MCPs together with 95\%-confidence intervals and the number of sent
messages. Dashes indicate that the MAPF-DP instance was not solved within a
runtime limit of 5 minutes. There is no obvious difference in the numbers of
sent messages of the 3 MAPF(-DP) solvers. However, AME seems to find MAPF-DP
plans with smaller average makespans than Adapted CBS, which seems to find
MAPF-DP plans with smaller average makespans than Push and Swap. The
approximate average makespans calculated by AME are underestimates but
reasonably close to the average makespans. AME and Push and Swap seem to run
faster than Adapted CBS. In fact, Adapted CBS did not solve MAPF-DP instances
with more than 35 agents within the runtime limit while AME and Push and Swap
seem to scale to larger numbers of agents than reported here (see also
Experiment 3).

\subsection{Experiment 2: Delay Probability Ranges}

We use AME with different delay probability ranges. We repeat Experiment 1
with 19 MAPF-DP instances generated from the MAPF-DP instance labeled ``random 1'' in
Experiment 1, one for each $\hat{t}_{max} = 2, 3 \ldots 20$. For each MAPF-DP
instance, the delay probabilities $p_i$ of all agents are sampled from the
delay probability range $(0, 1-1/\hat{t}_{max})$ by sampling the average
number of time steps $\hat{t}_i = 1/(1-p_i)$ needed for the successful
execution of single move actions uniformly at random from $(1,\hat{t}_{max})$
and then calculating $p_i = 1-1/\hat{t}_i$.

\begin{table}[t]
\tiny
\centering
\caption{Results of AME for MAPF-DP instances with 35 agents on a 30$\times$30
  4-neighbor grid with 10\% randomly blocked cells and different delay probability ranges $(0,
  1-\frac1{\hat{t}_{max}})$.}
\label{tab:2}
\begin{tabular}{ccccc}
\hline
\multicolumn{1}{c}{$\hat{t}_{max}$} & \multicolumn{1}{c}{runtime (s)} & \multicolumn{1}{c}{\begin{tabular}[c]{@{}c@{}}approximate average makespan\end{tabular}} & \multicolumn{1}{c}{\begin{tabular}[c]{@{}c@{}}average makespan\end{tabular}} & \multicolumn{1}{c}{messages} \\
\hline
2  & 0.073 & 77.92  & 84.30 $\pm$ 0.42  & 251 \\
3  & 0.525 & 123.92 & 131.12 $\pm$ 0.79 & 301 \\
4  & 0.356 & 144.61 & 157.88 $\pm$ 0.96 & 287 \\
5  & 0.311 & 133.55 & 157.00 $\pm$ 0.98 & 278 \\
6  & 0.623 & 168.51 & 192.76 $\pm$ 1.46 & 299 \\
7  & 0.346 & 264.78 & 279.51 $\pm$ 2.05 & 289 \\
8  & 0.236 & 333.09 & 349.72 $\pm$ 2.69 & 293 \\
9  & 0.779 & 260.58 & 271.71 $\pm$ 2.28 & 294 \\
10 & 1.751 & 307.63 & 336.95 $\pm$ 2.26 & 305 \\
11 & 2.528 & 337.15 & 375.46 $\pm$ 2.74 & 312 \\
12 & 1.374 & 323.87 & 383.25 $\pm$ 2.53 & 300 \\
13 & 0.683 & 381.63 & 413.18 $\pm$ 3.19 & 282 \\
14 & 2.583 & 440.94 & 498.30 $\pm$ 3.32 & 278 \\
15 & 1.414 & 470.06 & 524.94 $\pm$ 3.95 & 295 \\
16 & 7.072 & 554.32 & 607.20 $\pm$ 4.26 & 316 \\
17 & 2.116 & 451.32 & 570.15 $\pm$ 3.90 & 275 \\
18 & 3.410 & 763.44 & 782.40 $\pm$ 6.08 & 306 \\
19 & 5.708 & 462.71 & 666.42 $\pm$ 5.29 & 309 \\
20 & 7.812 & 490.26 & 591.35 $\pm$ 3.73 & 323 \\
\hline
\end{tabular}
\end{table}

\begin{figure}[t]
  \centering
  \includegraphics[width=0.75\columnwidth]{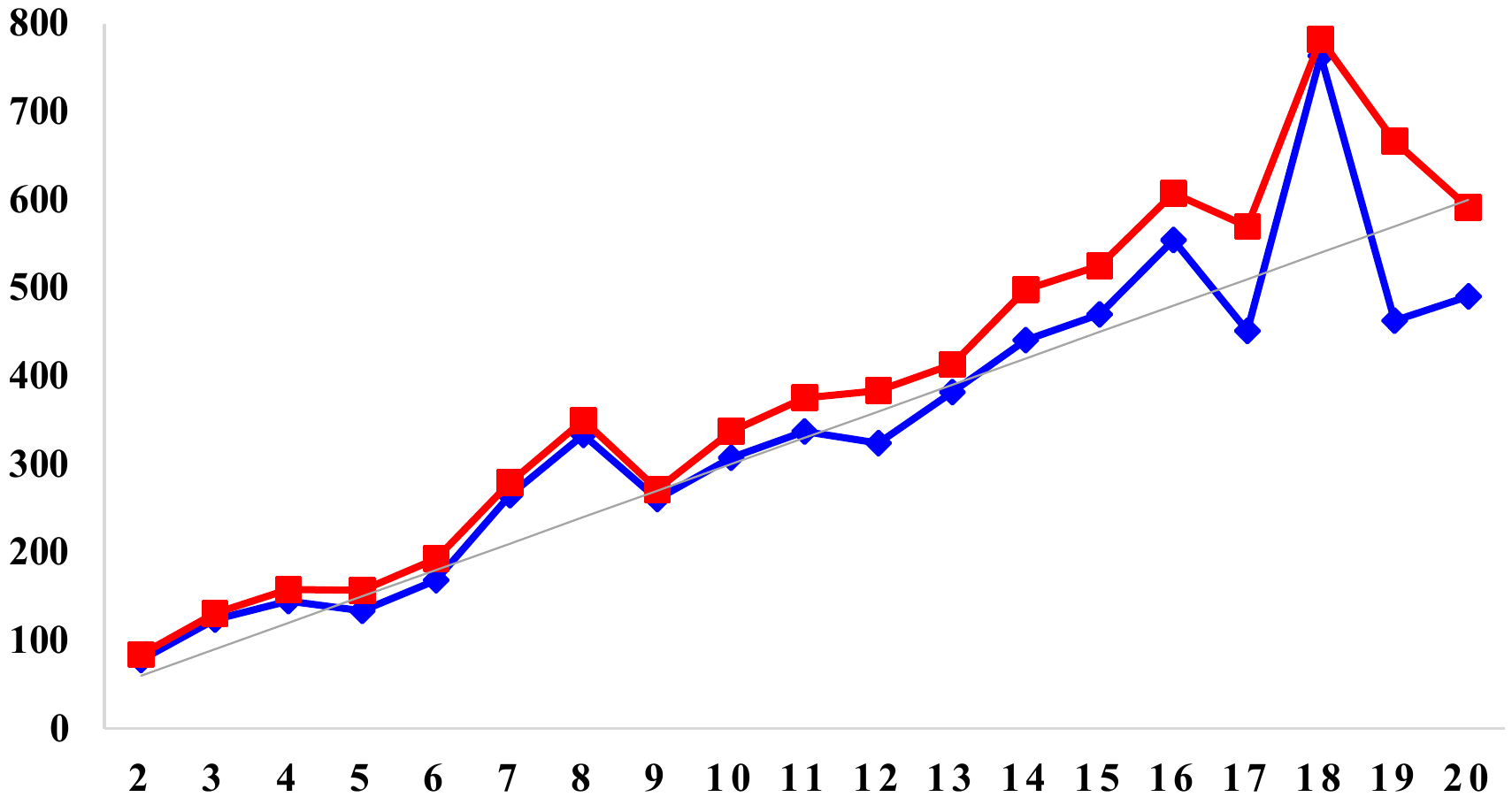}
  \caption{Visualization of Table~\ref{tab:2}, where the x-axis shows the
    average number of time steps $\hat{t}_{max}$ needed for the successful
    execution of single move actions. The average makespans are shown in red,
    and the approximate average makespans calculated by AME are shown in
    blue. The grey line corresponds to $30 \hat{t}_{max}$.}\label{fig:tab2}
\end{figure}

Table~\ref{tab:2} reports the same measures as used in Experiment 1, and
Figure~\ref{fig:tab2} visualizes the results. Larger delay probability ranges
seem to result in larger runtimes, approximate average makespans calculated by
AME and average makespans (although there is lots of
noise). The differences between the approximate average makespans calculated
by AME and average makespans are larger as well but remain reasonable.

\subsection{Experiment 3: Numbers of Agents}

We use AME with different numbers of agents. We repeat Experiment 1 with 50
MAPF-DP instances in 30$\times$30 4-neighbor grids generated as in Experiment 1 for each number of agents.

\begin{table}[t]
\tiny
\centering
\caption{Results of AME for MAPF-DP instances with different numbers of agents
  on 30$\times$30 4-neighbor grids with 10\% randomly blocked cells and
  delay probability range $(0, 1/2)$.}
\label{tab:3}
\begin{tabular}{cccccc}
\hline
\multicolumn{1}{c}{agents} & \multicolumn{1}{c}{solved (\%)} &
\multicolumn{1}{c}{runtime (s)} & \multicolumn{1}{c}{\begin{tabular}[c]{@{}c@{}}approximate\\average\\makespan\end{tabular}} & \multicolumn{1}{c}{\begin{tabular}[c]{@{}c@{}}average makespan\end{tabular}} & \multicolumn{1}{c}{\begin{tabular}[c]{@{}c@{}}messages\end{tabular}}\\
\hline
50  & 0.94 & 0.166   & 69.32 & 75.19 & 474.62  \\
100 & 0.68 & 4.668   & 78.48 & 87.29 & 1,554.71 \\
150 & 0.10  & 134.155 & 81.77 & 96.43 & 2,940.40 \\
200 & 0    & -       & -     & -     &         \\
\hline
\end{tabular}
\end{table}

Table~\ref{tab:3} reports the same measures as used in Experiment 1, averaged over all MAPF-DP
instances that were solved within a runtime limit of 5 minutes. AME solves
most MAPF-DP instances with 50 agents and then degrades gracefully with the
number of agents.

\subsection{Experiment 4: Plan-Execution Policies}

\begin{table}[t]
\Huge
\centering
\caption{Results of AME for the 18 solved MAPF-DP instances from Experiment 1 and different plan-execution policies.}
\label{tab:4}
\resizebox{\columnwidth}{!}{
\begin{tabular}{c|cc|cc|cc}
\hline
\multicolumn{1}{c|}{} & \multicolumn{2}{c|}{MCPs}
& \multicolumn{2}{c|}{FSPs}
& \multicolumn{2}{c}{\begin{tabular}[c]{@{}c@{}}Dummy\\Plan-Execution\\Policies\end{tabular}}                                                                                                                                     \\
\hline
\multicolumn{1}{c|}{id} &
\multicolumn{1}{c}{\begin{tabular}[c]{@{}c@{}}average\\makespan\end{tabular}} &
\multicolumn{1}{c|}{messages} &
\multicolumn{1}{c}{\begin{tabular}[c]{@{}c@{}}average\\makespan\end{tabular}} &
\multicolumn{1}{c|}{messages} &
\multicolumn{1}{c}{\begin{tabular}[c]{@{}c@{}}average\\makespan\end{tabular}} &
\multicolumn{1}{c}{\begin{tabular}[c]{@{}c@{}}average\\collisions\end{tabular}} \\
\hline
random 1  & 71.28 $\pm$ 0.34 & 267 & 140.29 $\pm$ 0.50 & 23,109 & 67.82 $\pm$ 0.35 & 16.68  \\
random 2  & 73.02 $\pm$ 0.29 & 257 & 143.55 $\pm$ 0.55 & 19,316 & 71.96 $\pm$ 0.31 & 14.27 \\
random 3  & 84.90 $\pm$ 0.40 & 373 & 160.43 $\pm$ 0.59 & 24,098 & 81.20 $\pm$ 0.37 & 27.71 \\
random 4  & 72.89 $\pm$ 0.37 & 251 & 141.71 $\pm$ 0.52 & 19,587 & 69.16 $\pm$ 0.36 & 25.38  \\
random 5  & 73.98 $\pm$ 0.31 & 255 & 141.49 $\pm$ 0.54 & 20,794 & 69.59 $\pm$ 0.32 & 14.98 \\
random 6  & 66.98 $\pm$ 0.36 & 257 & 115.98 $\pm$ 0.51 & 20,597 & 66.76 $\pm$ 0.37 & 15.19 \\
random 7  & 71.34 $\pm$ 0.36 & 269 & 124.03 $\pm$ 0.54 & 20,481 & 70.79 $\pm$ 0.38 & 16.53  \\
random 8  & 51.72 $\pm$ 0.35 & 164 & 96.04 $\pm$ 0.46  & 16,665 & 51.65 $\pm$ 0.38 & 8.81  \\
random 9  & 61.30 $\pm$ 0.27 & 247 & 113.76 $\pm$ 0.46 & 20,976 & 58.52 $\pm$ 0.23 & 10.33 \\
random 10 & 64.77 $\pm$ 0.38 & 234 & 114.04 $\pm$ 0.50 & 19,834 & 64.00 $\pm$ 0.38 & 17.51 \\
\hline
warehouse 1  & 124.18 $\pm$ 0.44 & 705 & 219.63 $\pm$ 0.65 & 28,794 & 122.42 $\pm$ 0.42 & 34.59 \\
warehouse 2  & 124.63 $\pm$ 0.51 & 762 & 235.35 $\pm$ 0.72 & 34,154 & 124.40 $\pm$ 0.60 & 68.68 \\
warehouse 3  & 117.00 $\pm$ 0.53 & 609 & 206.29 $\pm$ 0.65 & 26,647 & 117.89 $\pm$ 0.54 & 29.61  \\
warehouse 4  & 117.31 $\pm$ 0.52 & 541 & 194.07 $\pm$ 0.59 & 24,889 & 116.02 $\pm$ 0.53 & 28.09 \\
warehouse 6  & 131.10 $\pm$ 0.59 & 710 & 205.54 $\pm$ 0.71 & 29,462 & 131.54 $\pm$ 0.60 & 37.41 \\
warehouse 7  & 96.54 $\pm$ 0.34  & 488 & 187.90 $\pm$ 0.59 & 22,401 & 95.80 $\pm$ 0.35  & 24.91 \\
warehouse 9  & 107.33 $\pm$ 0.42 & 462 & 187.80 $\pm$ 0.56 & 18,950 & 105.63 $\pm$ 0.45 & 22.21 \\
warehouse 10 & 127.36 $\pm$ 0.53 & 909 & 226.95 $\pm$ 0.73 & 32,903 & 127.59 $\pm$ 0.55 & 43.78  \\
\hline
\end{tabular}
}
\end{table}

We use AME with 3 plan-execution policies, namely 1) MCPs, 2) FSPs and 3)
dummy (non-robust) plan-execution policies that always provide GO commands. We
repeat Experiment 1 for each plan-execution policy.

Table~\ref{tab:4} reports for each solved MAPF-DP instance and plan-execution policy
the average makespan over 1,000 plan-execution runs together with
95\%-confidence intervals, the number of sent messages for MCPs and FSPs and
the average number of collisions for dummy plan-execution policies. The number
of sent messages is zero (and thus not shown) for dummy plan-execution
policies since, different from MCPs and FSPs, they do not prevent
collisions. The average makespan for MCPs seems to be only slightly larger
than that for dummy plan-execution policies, and the average makespan and
number of sent messages for MCPs seem to be smaller than those for FSPs.

\section{Conclusions}

In this paper, we formalized the Multi-Agent Path-Finding Problem with Delay
Probabilities (MAPF-DP) to account for imperfect plan execution and then
developed an efficient way of solving it with small average makespans, namely
with Approximate Minimization in Expectation (a 2-level MAPF-DP solver for
generating valid MAPF-DP plans) and Minimal Communication Policies
(decentralized robust plan-execution policies for executing valid MAPF-DP
plans without collisions).

\small
\bibliographystyle{aaai}
\bibliography{references}
\end{document}